\documentclass[10pt,journal,compsoc]{IEEEtran}

\usepackage{cite}

\ifCLASSINFOpdf
   \usepackage[pdftex]{graphicx}
\else
   \usepackage[dvips]{graphicx}
\fi

\usepackage{amsmath}

\ifCLASSOPTIONcompsoc
 \usepackage[caption=false,font=normalsize,labelfont=sf,textfont=sf]{subfig}
\else
 \usepackage[caption=false,font=footnotesize]{subfig}
\fi

\usepackage{url}
\usepackage{makecell}
\usepackage{amssymb}
\usepackage{multirow}
\usepackage{booktabs}
\usepackage{float}

\begin{document}

\title{Dependency Learning for Legal Judgment Prediction with a Unified Text-to-Text Transformer}

\author{Yunyun~Huang*,~%
		Xiaoyu~Shen*,~%
		Chuanyi~Li,~
		Jidong~Ge,~
		Bin Luo
\IEEEcompsocitemizethanks{
\IEEEcompsocthanksitem Y. Huang, C. Li, J. Ge, and B. Luo are with State Key Laboratory for Novel Software Technology, Software Institute, Nanjing University, Nanjing 210093, China.
E-mails: oli\_yun@163.com, lcy@nju.edu.cn, gjd@nju.edu.cn, luobin@nju.edu.cn
\IEEEcompsocthanksitem X. Shen is with Amazon Alexa AI, Berlin, Germany (Work done before joining). 
E-mail: gyouu@amazon.com
\IEEEcompsocthanksitem * Authors contributed equally.
}
}

\markboth{}%
{Shell \MakeLowercase{\textit{et al.}}: Bare Demo of IEEEtran.cls for Computer Society Journals}

\IEEEtitleabstractindextext{%
\begin{abstract}
Given the fact of a case, \textbf{L}egal \textbf{J}udgment \textbf{P}rediction (LJP) involves a series of sub-tasks such as predicting violated law articles, charges and term of penalty. We propose leveraging a \emph{unified text-to-text Transformer} for LJP, where the dependencies among sub-tasks can be naturally established within the auto-regressive decoder. Compared with previous works, it has three advantages: (1) it fits in the pretraining pattern of masked language models, and thereby can benefit from the semantic 
prompts of each sub-task rather than treating them as atomic labels, (2) it utilizes a single unified architecture, enabling full parameter sharing across all sub-tasks, and (3) it can incorporate both classification and generative sub-tasks. We show that this unified transformer, albeit pretrained on general-domain text, outperforms pretrained models tailored specifically for the legal domain. Through an extensive set of experiments, we find that the best order to capture dependencies is \emph{different} from human intuitions, and the most reasonable logical order for humans can be sub-optimal for the model. We further include two more auxiliary tasks: court view generation and article content prediction, showing they can not only improve the prediction accuracy, but also provide interpretable explanations for model outputs even when an error is made.  With the best configuration, our model outperforms both previous SOTA and a single-tasked version of the unified transformer by a large margin. Code and dataset are available at \url{https://github.com/oli-yun/Dependency-LJP}.
\end{abstract}

}

\maketitle

\IEEEdisplaynontitleabstractindextext

\IEEEpeerreviewmaketitle

\section{Introduction}

\IEEEPARstart{L}egal \textbf{J}udgment \textbf{P}rediction (LJP) aims to predict the judgment results of legal cases according to the fact descriptions, which involves multiple sub-tasks depending on country-specific standards. Under the Civil Law system, these sub-tasks usually include (1) finding the violated law articles, (2) defining the charge, and (3) deciding the term of penalty. 
Automating these sub-tasks is of great interest in that it can not only improve the working efficiency of judges and lawyers, but also provide basic legal assistance to non-professionals.

Earlier works solved these sub-tasks as independent text classification problems, while ignoring the close dependencies among different sub-tasks~\cite{ulmer_quantitative_1963,keown_mathematical_1980,liu_exploring_2006,lin_exploiting_2012,li_markov_2018}. Recently, many works have demonstrated benefits of modelling such dependencies. The simplest way is a pipeline method that conditions each sub-task on the prediction of its dependent sub-tasks~\cite{luo_learning_2017,long_automatic_2019,wei_external_2019,chen_charge-based_2019}. However, it requires an independent sub-module for each sub-task, which complicates the overall system and prevents information sharing across sub-tasks. Another line of work is multi-tasking by sharing the same model parameters among all sub-tasks~\cite{zhong_iteratively_2020,xu_distinguish_2020}. Nonetheless, the prediction of each sub-task is still based solely on the fact description. The dependency learning is only implicitly reflected through parameter sharing while ignoring the logical orders of sub-tasks. A few works enable both parameter sharing and dependency learning under specified logical orders~\cite{zhong_legal_2018,yang_legal_2019,zhong2020does}. They design different architectures for each purpose, e.g., CNN for fact encoding and LSTM for dependency learning, such that all sub-tasks can share the same fact encoder while maintaining their own task-specific representations. 

We argue that this way of combining different architectures is by no means the optimal solution, and instead propose leveraging a unified text-to-text Transformer for dependency learning in LJP. Specifically, we implement our model based on the pretrained T5~\cite{raffel2020exploring} architecture, where all sub-tasks are considered as ``masked spans" from the original legal judgment document. In the training stage, the model learns all sub-tasks by reconstructing ``masked spans" sequentially. The generated sequence can be easily mapped to classification labels for the evaluation purpose. Compared with previous works, it has the following advantages: (1) It fits in the same pattern as the pretraining objective of the T5 model. The structure of the legal judgment documents can naturally serve as semantic prompts~\cite{schick2021s,tam2021improving} for different sub-tasks, which provides the model with a much better initialization especially under the fewshot scenario. (2) There will be no separate architectures or task-specific representations. It utilizes a single unified Transformer, enabling full parameter sharing across all sub-tasks. (3) It applies to both classification and generative tasks, so that there will be no limitation on the sub-task types.

Experiments on real-world criminal cases show that our proposed unified architecture, albeit pretrained on general-domain text, outperforms pretrained models tailored specifically for the legal domain. It performs especially well on the Macro-F1 metric, implying a strong generalization on low-frequency labels. In contrast, all previous works overfit to high-frequency labels without exception. It also demonstrates strong fewshot learning capability with only hundreds of training examples.
\IEEEpubidadjcol
We conducted an extensive set of experiments on the effects of sub-task orders under dependency learning. We find that the best order is \emph{different} from human intuitions, and the most reasonable logical order for humans can be sub-optimal for the model. The model is seeking a sweet spot to balance well the dependency on other sub-tasks and error propagation. When the risk of error propagation outweighs the benefit from dependency learning, it prefers putting ahead the sub-task, even if it should be solved in a later logical order for humans.

Apart from the three commonly used sub-tasks, we introduce two more auxiliary tasks: court view generation and law article content prediction. For the former, we show it is able to not only improve the prediction accuracy, but also provide interpretable explanations to justify the logic behind, even when the model makes a wrong prediction. For the latter, we find that it boosts the prediction accuracy when combining with every single task. In the law article prediction sub-task, it even outperforms the text-matching model which learns a full interaction between facts and law article contents, while being hundreds of times faster.

\begin{figure}[t]
\centering
\includegraphics[width=.8\linewidth]{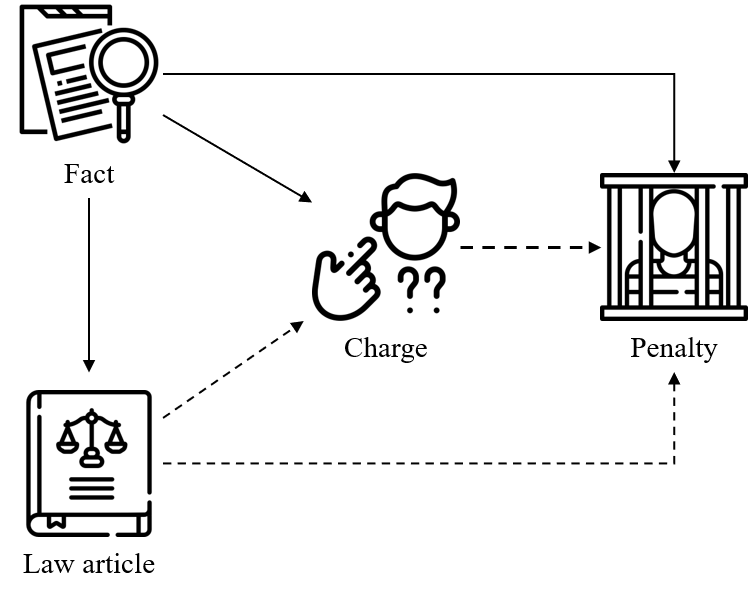}
\caption{\small Process of LJP. Lines indicate dependency relations.}
\label{fig2}
\end{figure}

In short, this work makes the following contributions:
\begin{itemize}
\item We propose leveraging a unified text-to-text Transformer for dependency learning in LJP. We show it significantly outperforms previous SOTA with a much stronger generalization on low-frequency labels, and performs decently even under the fewshot scenario.
\item We perform an extensive set of experiments to find the optimal order for dependency modelling. We find the best order for the model depends on the trade-off between sub-task dependency and error propagation, and can be different from human intuitions.
\item We introduce two new auxiliary tasks: court view generation and law article content prediction. They can further enhance the prediction accuracy, while providing interpretable explanations to the model outputs.
\end{itemize}

\begin{figure*}[ht]
\centering
\includegraphics[width=\textwidth]{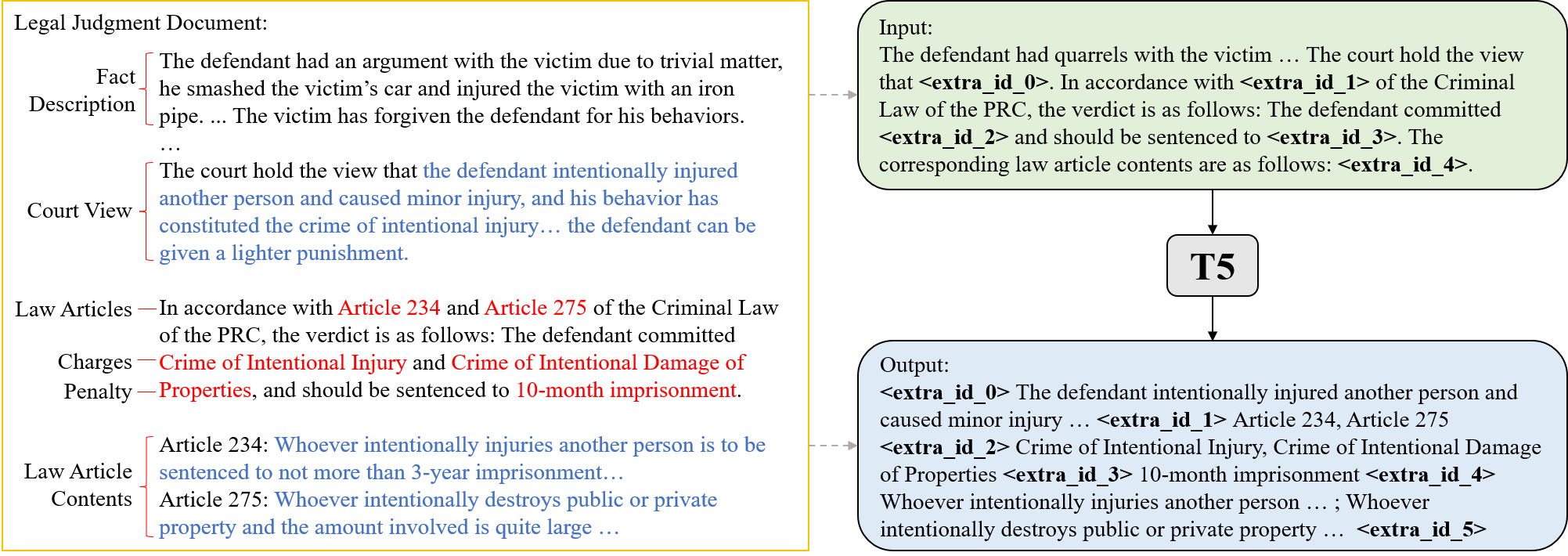}
\caption{\small Example of processing the legal judgment document to get the input-output formats. Spans in red are main tasks in LJP and spans in blue are auxiliary tasks. We mask colored words from the original document as the input. The output is a sequence of masked spans. The order of spans in the output can be adjusted to model different dependency relations.}
\label{fig1}
\end{figure*}

\section{Problem Formulation}
We focus on LJP tasks under the civil law system, but similar techniques can be easily extended to other systems as well. Given factual descriptions of one case, LJP under the civil law system involves a set of sub-tasks, including predicting the violated law articles, charges and term of penalty. The violated law article is legal basis for judgment results and charge is the category of the committed crime, e.g., theft, intentional injury, etc. We illustrate the process of LJP in Figure~\ref{fig2}. For human judges, there exist dependencies with a strict order among these sub-tasks. A judge will always first decide the violated law articles, then determine the charges according to the instructions of relevant law articles. Finally, the term of penalty will be confirmed based on these.

The violated law article and charge have a pre-defined fixed set of possible labels, but the term of penalty can be rather diverse across different crimes. Following common practices, we simplify it into three main types: fixed-term imprisonment, life imprisonment and death penalty. For fixed-term imprisonment, the model needs to predict the exact length in the unit of month. Since it is stipulated that the maximum period of fixed-term imprisonment is 15 years and that of the criminal who commits several crimes is 25 years, we set the value of life imprisonment as $350$ and death penalty as $400$ in accordance with \cite{zhong_legal_2018}.

Formally speaking, we denote our dataset with $n$ cases as $D=(F^i, \{A_1^i,\dots,A_k^i\}, \{C_1^i,\dots,C_l^i\}, P^i)_{i=1}^n$ where each case $D^i$ consists of a fact description $F^i$, a set of $k$ violated law articles $\{A_1^i,\dots,A_k^i\}$, a set of $l$ accused charges $\{C_1^i,\dots,C_l^i\}$ and a penalty term $P^i$. All of these items are composed of a sequence of words.
Law articles and charges have their corresponding sets of categories and each law article has its own textual definition. Our goal is to learn a classifier $\zeta$ which is able to predict the judgment results for a given fact, i.e., $(\{A_1^i,\dots,A_k^i\}, \{C_1^i,\dots,C_l^i\}, P^i)=\zeta (F^i)$.

\section{Methods}
In this section, we first introduce the T5 model and semantic prompts, then describe how we can leverage the existing legal judgment documents to construct semantic prompts for LJP based on T5. Finally, we explain how it enables easy dependency learning for arbitrary sub-tasks.
\subsection{Introduction of T5} We construct our model based on T5~\cite{raffel2020exploring}, a large-scaled pretrained language model (PLM) based on the Transformer architecture~\cite{vaswani2017attention}. At each time step $t$, it uses a left-to-right language model $\mathcal{D}$ to predict $\mathbf{y}_t$ conditioned on previous generated tokens and hidden states $\mathbf{h}$ from a separate self-attention-based encoder $\mathcal{E}$ for input $\mathbf{x}$:
$$
\mathbf{h} = \mathcal{E}(\mathbf{x}) \quad \mathbf{y}_t = \mathcal{D}(\mathbf{h}, \mathbf{y}_1, \dots, \mathbf{y}_{t-1})
$$
T5 unifies the training framework of natural language understanding and natural language generation into the text-to-text form. It is pretrained with the span-corruption objective, where consecutive spans of input tokens are replaced with a mask token and the model is trained to reconstruct the masked-out tokens. During pretraining, the input is the masked sentence and the target is a sequence of masked spans where each span starts with its indicator (e.g., $<$extra\_id\_0$>$). The mean length of a span is 3 and 15\% of the original text sequence are masked out.
\subsection{Semantic prompts} Applying semantic prompts to PLMs has become increasingly popular. Under this paradigm, downstream tasks are reformulated to look more
like those solved during the original PLM pretraining objective with the help of a textual prompt. For example, when recognizing the emotion of a social media post, ``I missed the bus today.", we may continue with a prompt “I felt so \underline{\hspace*{0.5cm}}”, and
ask the LM to fill the blank with an emotion-bearing word~\cite{liu2021pre}. The filled word can be mapped to emotion labels by pre-defined rules. It has shown consistent improvement over the traditional supervised finetuning especially in the fewshot scenario~\cite{brown2020language,schick2021s} because it can map tasks into existing semantic knowledge inside the PLM instead of learning from scratch.
\subsection{Prompts For LJP} It is quite straightforward to construct prompts for LJP. Since all sub-task labels are themselves extracted from legal judgment documents by regex rules~\cite{xiao_cail2018_2018}, we can directly leverage the structure of the document as prompts for different sub-tasks. A legal judgment document starts with a fact, following by a set of judgment results and explanations. We can mask out the words describing the judgment results from the document the same regex rule as used in ~\cite{xiao_cail2018_2018}, input this corrupted document into the T5 model, then ask the model to predict original spans in order. An illustration is shown in Figure~\ref{fig1}. By this means, the training shares exactly the same format as the original pretraining objective of T5, such that we can make the most use of the knowledge inside the pretrained model. 
After the model predicts the sequence, we can convert it into sub-task labels with a one-to-one mapping.
\subsection{Dependency Learning}
The dependency learning across sub-tasks is naturally established with the autoregressive decoder from the T5 model, where the prediction of one span conditions on all the previously decoded spans. Taking predicting law articles and term of penalty as example, we suppose that the latter depends on the former here. We first use a prompting function to modify the input fact $F^i$ into a prompt: 
$$\tilde{F}^i=f_{prompt}(F^i)$$ 
and the output is the concatenation of different sub-tasks labels which are connected by their indicators:
$$Y^i=\{I_1;A_1^i;\dots;A_k^i;I_2;P_i;I_3\}$$ 
where $I_k$ represents the indicator of the $k$-th sub-task and the last one is the end signal. Then we pass $\tilde{F}^i$ into T5 and get output tokens step by step. When the model decode $I_2$ at time step $t$, we can use previous generated tokens $\{\mathbf{y}_1, \dots, \mathbf{y}_{t-1}\}$ to map corresponding law articles $\{A_1^i,\dots,A_k^i\}$, and these tokens are also used to predict term of penalty which is equivalent to: 
$$
P^i=\mathcal{D}(\mathcal{E}(\tilde{F}^i), A_1^i, \dots, A_k^i)
$$
In this way, we build dependencies between different sub-tasks. 
Same as in text generation tasks, the decoder conditions on the ground-truth spans during training, and on the self-predicted spans during inference. The order of spans in the decoding side reflects the dependency relations. We can simply adjust the order of decoded spans (together with its indicator) to control the logical order of sub-tasks.

\begin{figure*}
    \centering
    \subfloat[Single-task Learning]{
    \includegraphics[width=.25\textwidth]{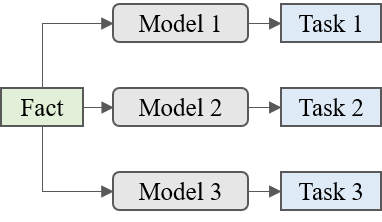}
    \label{fig:stl}}
    \hspace{.75cm}
    \subfloat[Multi-task Learning]{
    \includegraphics[width=.25\textwidth]{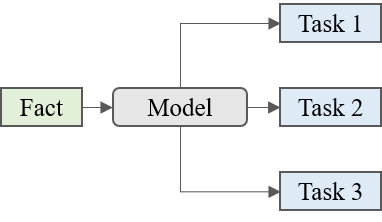}
    \label{fig:mtl}}
    \hspace{.75cm}
    \subfloat[Dependency Learning]{
    \includegraphics[width=.285\textwidth]{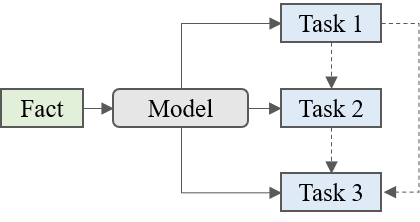}
    \label{fig:dl}}
    \caption{\small Three types of learning methods used in LJP. Single-task learning applies independent models for different tasks. Multi-task learning applies a single model for all tasks, but there is no dependency across tasks. Dependency-learning applies a single model to capture all dependencies within tasks.}
    \label{fig:learning_method}
\end{figure*}

\subsection{Auxiliary Tasks} The above framework provides us a unified way to incorporate arbitrary sub-tasks existing in the legal judgment document. We can easily add on more tasks by masking out the corresponding spans from the original document. In this work, we experiment with two more auxiliary tasks: court view generation and law article content explanation. Court view can be regarded as the interpretation for the sentence of a case. It summarizes the fact description and specifies related charges,
then explains the rationale to derive the judgment. Law article content includes a complete description of the premises for violation and the scope of sentence. Both of them can be useful for LJP since they provide the background basis and explanations on how the judgment results are derived. By co-training with these two auxiliary tasks, the model is able to ``explain" its own predictions by generating the corresponding court view and article content, which is crucial for a trustworthy LJP system.

\section{Experiments}
We conduct our experiments on a dataset collected from the China Judgments Online~\footnote{\url{https://wenshu.court.gov.cn/}}. In this section, we first explain how we collect such dataset, then go over a set of baseline systems we compare with, and finally present the findings.

\subsection{Dataset}
In the field of LJP, existing works often experimented on the CAIL dataset~\cite{xiao_cail2018_2018} or extracted task-specific texts from the published judgment documents to construct their own dataset~\cite{luo_learning_2017, ye_interpretable_2018, zhong_iteratively_2020}. However, none of them contains all the 5 tasks we need. Furthermore, these datasets have been through a series of pre-processing which makes it difficult to map each case to the original legal judgment document. Therefore, we construct a dataset ourselves following the same mechanism as in ~\cite{xiao_cail2018_2018}. First of all, we filtered those cases involving multiple defendants because different defendants correspond to different trial results. Besides, the top 102 law articles in Chinese Criminal Law are not relevant to specific charges, we also filtered these labels. In this way, we got a dataset involving 200 charges and 183 law articles from 225,843 criminal judgment documents. We randomly selected 12,810 cases to form the test set and ensured it covered all types of charges and law articles. We also selected 12,634 cases to construct our validation set and the rest is training set. 
Each case refers to 1.14 law articles and involves 1.06 charges on average, but has strictly one term of penalty.
We provide more detail statistics in Table~\ref{tab:statistic}.

\begin{table}[t]
    \caption{Statistics of our dataset. We show the number of unique facts, articles, etc. and the average number of words in each of them.}
    \centering
    \resizebox{\linewidth}{!}{
    \begin{tabular}{c|r|r|r|r|r|r}
    \toprule
        & Fact & Article & Charge & Penalty & View & Content \\
        \midrule
        \# Unique & 225,843 & 183 & 200 & 207 & 225,843 & 183 \\
        \# Words & 199.46 & 3 & 7.26 & 6.41 & 148.28 & 137.9 \\
    \bottomrule
    \end{tabular}}
    \label{tab:statistic}
\end{table}

\subsection{Experimental Settings}
\subsubsection{Model configuration} We fine-tune our model based on mT5 base~\cite{xue2021mt5}, a multilingual variant of T5 that can be used to fine-tune on Chinese tasks. For training, we set batch size as 128 and adopt the gradient accumulation strategy. All models are trained for a maximum of 30 epochs with early-stopping, and the model which performs best on the validation set will be selected. Max length of input and output are both set to 512. 
We use the AdaFactor optimizer~\cite{shazeer2018adafactor} which can relieve memory pressure to a certain extent.

\subsubsection{Metrics} For law article and charge, we employ micro-F1 and macro-F1 as evaluation metrics. Since penalty is regarded as regression task, we calculate the log-distance between prediction and ground-truth following~\cite{zhong_how_2020}, a smaller log-distance indicates better prediction.

\subsubsection{Baselines}
We compare with the following baselines:
\begin{itemize}
\item \textbf{LSTM}~\cite{hochreiter1997long}: We employ two-layer bidirectional LSTM to encode factual description and then make prediction for each task with full-connected neural network.
\item \textbf{CNN}~\cite{kim2014conv}: We employ CNN with multiple filter widths as fact encoder and then make prediction.
\item \textbf{FactLaw}~\cite{luo_learning_2017}: It predicts top-k law articles with SVM, then encodes fact and law article contents with HAN~\cite{yang2016hierarchical} and predicts the result.
\item \textbf{TopJudge}~\cite{zhong_legal_2018}: It encodes facts with CNN, and then models task dependencies with LSTM. 
\item \textbf{LBERT}~\cite{zhong_open_2019}: It pretrained BERT on a large scale number of Chinese criminal judgment documents which shows better performance than original BERT~\cite{devlin2019bert} in LJP. We use this model to encode fact following by three different fully connected networks for prediction of main tasks respectively.
\end{itemize}

\begin{table*}[t]
\caption{Main results of different models trained on 10,000 samples.}
\centering
\resizebox{.95\linewidth}{!}{
\begin{tabular}{c |l |c c|c c|c|c c|c c|c}
    \toprule
     & & \multicolumn{5}{c|}{Dev} & \multicolumn{5}{c}{Test} \\
    \midrule
     & & \multicolumn{2}{c}{Article} & \multicolumn{2}{|c|}{Charge} & Penalty & \multicolumn{2}{c}{Article} & \multicolumn{2}{|c|}{Charge} & Penalty \\
    \midrule
    Method & \makecell[c]{Model} & MiF & MaF & MiF & MaF & Dis $\downarrow$ & MiF & MaF & MiF & MaF & Dis $\downarrow$ \\
    \midrule
    \multirow{5}{*}{Single-task Learning} & STL-LSTM & 0.697 & 0.304 & 0.709 & 0.293 & 2.170 & 0.688 & 0.250 & 0.706 & 0.243 & 2.178 \\
    & STL-CNN & 0.730 & 0.330 & 0.730 & 0.315 & 2.177 & 0.718 & 0.275 & 0.717 & 0.261 & 2.188 \\
    & STL-TopJudge & 0.760  & 0.353 & 0.775 & 0.338 & 2.139 & 0.753 & 0.315 & 0.777 & 0.330  & 2.097 \\
    & STL-LBERT & 0.817 & 0.438 & 0.803 & 0.428 & 2.053 & 0.809 & 0.394 & 0.800 & 0.386 & 2.063 \\
    & STL-T5 & 0.763 & 0.488 & 0.818 & 0.550 & 1.980 & 0.757 & 0.429 & 0.807 & 0.499 & 1.969 \\
    \midrule
    \multirow{5}{*}{Multi-task Learning} & MTL-LSTM & 0.719 & 0.320 & 0.720 & 0.296 & 2.185 & 0.708 & 0.265 & 0.708 & 0.247 & 2.185 \\
    & MTL-CNN & 0.736 & 0.345 & 0.740 & 0.323 & 2.183 & 0.719 & 0.286 & 0.725 & 0.274 & 2.201 \\
    & MTL-TopJudge & 0.774 & 0.365 & 0.777 & 0.339 & 2.093 & 0.771 & 0.313 & 0.767 & 0.289 & 2.110 \\
    & MTL-LBERT & 0.805 & 0.418 & 0.810 & 0.418 & 2.115 & 0.801 & 0.380 & 0.801 & 0.372 & 2.123 \\
    & MTL-T5 & 0.813 & 0.483 & 0.826 & 0.506 & 1.936 & 0.808 & 0.431 & 0.814 & 0.455 & 1.927 \\
    \midrule
    \multirow{3}{*}{Dependency Learning }& FactLaw & 0.733 & 0.416 & 0.535 & 0.220 & 2.087 & 0.717 & 0.403 & 0.536 & 0.217 & 2.096 \\
    & Dependent-TopJudge & 0.777 & 0.381 & 0.787 & 0.393 & 2.047 & 0.772 & 0.327 & 0.781 & 0.335 & 2.054 \\
    & Dependent-T5$^{\bigstar}$ & 0.825 & 0.526 & 0.836 & 0.558 & 1.877 & 0.816 & 0.469 & 0.824 & 0.512 & 1.878 \\
    \midrule
    \multirow{3}{*}{+ Auxiliary Tasks} & $\bigstar$ + View & 0.827 & 0.538 & \textbf{0.843} & \textbf{0.581} & 1.854 & \textbf{0.823} & \textbf{0.498} & \textbf{0.839} & \textbf{0.530} & 1.846 \\
    & $\bigstar$ + Content & \textbf{0.828} & 0.532 & 0.838 & 0.543 & \textbf{1.842} & 0.821 & 0.486 & 0.826 & 0.515 & 1.844 \\
    & $\bigstar$ + View \& Content & 0.827 & \textbf{0.543} & 0.841 & 0.567 & 1.867 & 0.819 & 0.481 & 0.830 & 0.521 & \textbf{1.840} \\
    \bottomrule
\end{tabular}}
\label{tab:res}
\end{table*}

FactLaw did not release the code, we set $k=5$ and reproduce it ourselves. Other baselines are implemented with the open-sourced code in~\cite{zhong_how_2020}. For all above models, we compare them under the following settings:
\begin{itemize}
\item \textbf{Single-task Learning (STL)}: As shown in Figure~\ref{fig:stl}, it applies independent models for different tasks. There is not any correlation among tasks.
\item \textbf{Multi-task Learning (MTL)}: As shown in Figure~\ref{fig:mtl}, it trains all tasks with a single model with parameter sharing, but there is no dependency across tasks. 
\item \textbf{Dependency Learning}: As shown in Figure~\ref{fig:dl}, it trains all tasks with parameter sharing plus explicit dependency modelling. Only FactLaw and TopJudge can apply to this setting.   
\end{itemize}

\subsection{Main Results}
We show the main comparison results in Table~\ref{tab:res}. All models are trained on the same 10,000 subset sampled from the whole training data as a pilot study. For the dependency-learning models, we use the order of article-charge-penalty for model prediction, which, as explained, is the order that humans follow to make the judgment results. For the experiments with auxiliary tasks, we append the auxiliary tasks in the end of the three main tasks.

As for the model architecture, pretrained \emph{Lbert and T5 have a clear advantage} compared with other models that are trained from scratch. T5 also significantly outperforms LBert, even though it has never been tailored for legal-specific text. T5 performs especially well on the Macro-F1 metric, with a consistent lead of $8\sim30\%$, suggesting it is able to generalize to low-frequency labels effectively, while all other models overfit only to high-frequency labels. For all architectures, \emph{MTL does not lead to significant improvement over STL} on most tasks, suggesting parameter sharing only is not an effective way for dependency learning.

When applying dependency learning, both the performance of TopJudge and T5 get improved. However, the improvement of Dependent-T5 over STL/MTL-T5 is much larger than that of Dependent-TopJudge over STL/MTL-TopJudge, indicating \emph{our proposed structure is a more efficient way to enable dependency learning than TopJudge}.

By adding the two auxiliary tasks individually, the performance of Dependency-T5 is again improved. However, combining both of them did not help further. The benefit that could get from auxiliary tasks could have been saturated.

\begin{figure*}[t]
\centering
\includegraphics[width=\textwidth]{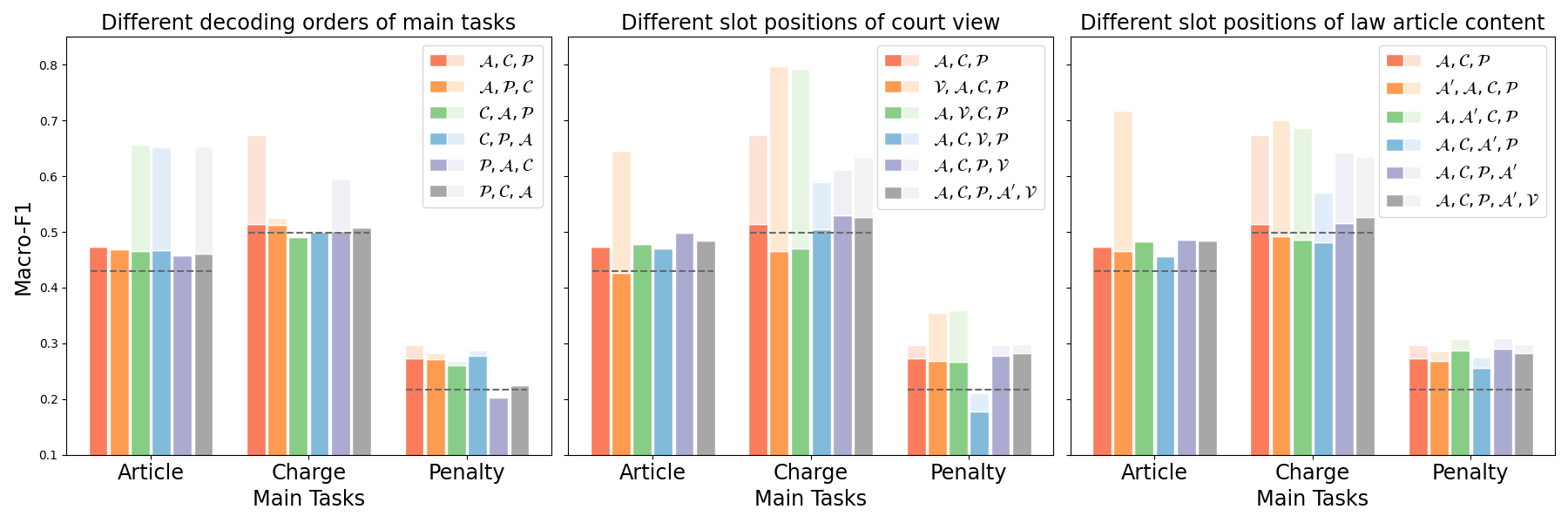}
\caption{\small Effects of decoding orders. For visualization, we divided the term of penalty into 11 categories to calculate Macro-F1 too. Bars in different colors indicate different orders of decoding article ($\mathcal{A}$), charge ($\mathcal{C}$), penalty ($\mathcal{P}$), court view ($\mathcal{V}$) and article content ($\mathcal{A'}$). Upper bars in lighter colors are results of making prediction based on the ground-truths of predecessor tasks. Dashed lines indicate single-task results.}
\label{fig3}
\end{figure*}

\subsection{Analysis}

\subsubsection{Orders and Dependencies}
We show the effects of decoding orders in Figure~\ref{fig3}, where we modify the orders of three main tasks and insert two auxiliary tasks into different positions. As can be observed, \emph{dependency learning benefits not only successor tasks, but also predecessor tasks}. Law article prediction under all the 6 order combinations outperform the single-task performance, suggesting \emph{the unified dependency learning framework provides a more effective way of parameter sharing than multi-task learning}. Even for the first decoding task that has no dependency on other tasks, it can still outperform multi-task learning under the same setting.

The prediction of article and charge correlate more with each other, and knowing one of them helps significantly the prediction of the other. Penalty can benefit more from article/charge than the other way around. This coincides with humans that the decision of penalty should be left at the end.

\emph{The best decoding order for each task can be different from human intuitions}. The reason is that there exists error propagation for machines. When the downside of error propagation outweighs the dependency of other tasks, machines will prefer putting ahead the task to maximize the performance. For example, humans tend to decide the penalty after confirming the charge, yet the model prefers predicting the penalty before the charge because the error propagation from charges will negatively impact the prediction.

This is more obvious for auxiliary tasks. For example, humans will first come with the court view because it includes the derivation that we need to decide the penalty. For the model, however, putting the court view before the penalty significantly deteriorates the results because the court view is long and prone to error propagation. The same holds for law article contents. \emph{For both auxiliary tasks, the best order is to put them at the end} so that they will not incur any error propagation for the main tasks.

\begin{table}[t]
    \caption{Comparison of leveraging law article contents for article prediction (all models are based on mT5). Dependent-T5 is more effective than text matching while being significantly faster.}
    \centering
    \resizebox{\linewidth}{!}{
    \begin{tabular}{l|c|c|r}
        \toprule
        Methods & MiF & MaF & Speed/s \\
        \midrule
        STL (article) & 0.757 & 0.429 & 0.37 \\
        Text Matching & 0.770 & 0.431 & 64.74 \\
        Dependent (article + content) & 0.817 & 0.488 & 0.37 \\
        \bottomrule
    \end{tabular}
    }
    \label{tab:content}
\end{table}

\subsubsection{Law Article Content}
The law article item and content have a strictly one-to-one relation. A more common way to leverage the context for article prediction is by text matching, where we turn it from multi-label classification to multiple binary classification problems. 
Following the common practice for text matching, we concatenate the fact with every single law article content as the input of T5, and decode a binary relevant/irrelevant output. We take the positive-negative ratio as 1:31 to construct training set and train it with cross-entropy loss following \cite{karpukhin2020dense}. When testing, the classification threshold is set as 0.7 (tuned to maximized the micro-F1). We compare it with the STL-T5 (predict article only) and Dependent-T5 (decode article then content). Results are shown in Table~\ref{tab:content}. We can see that Dependent-T5 outperforms the text-matching method without affecting the inference speed because the content prediction is only used in training. Text matching, on the contrary, slows down the inference significantly as it needs to match over all articles one by one. Even with parallel batching, it still doubles the time with hundreds more memory cost.

\begin{figure}[t]
    \centering
    \includegraphics[width=.9\linewidth]{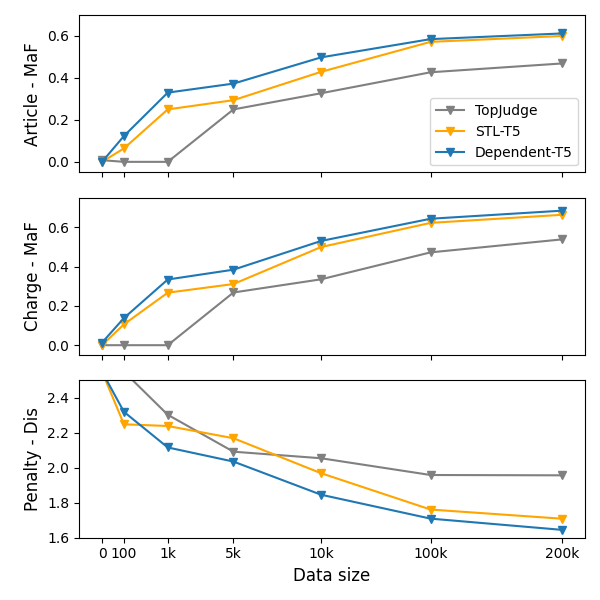}
    \caption{\small Performance by training on varying data sizes.}
    \label{fig:data_size}
\end{figure}

\begin{figure*}
    \includegraphics[width=.95\textwidth]{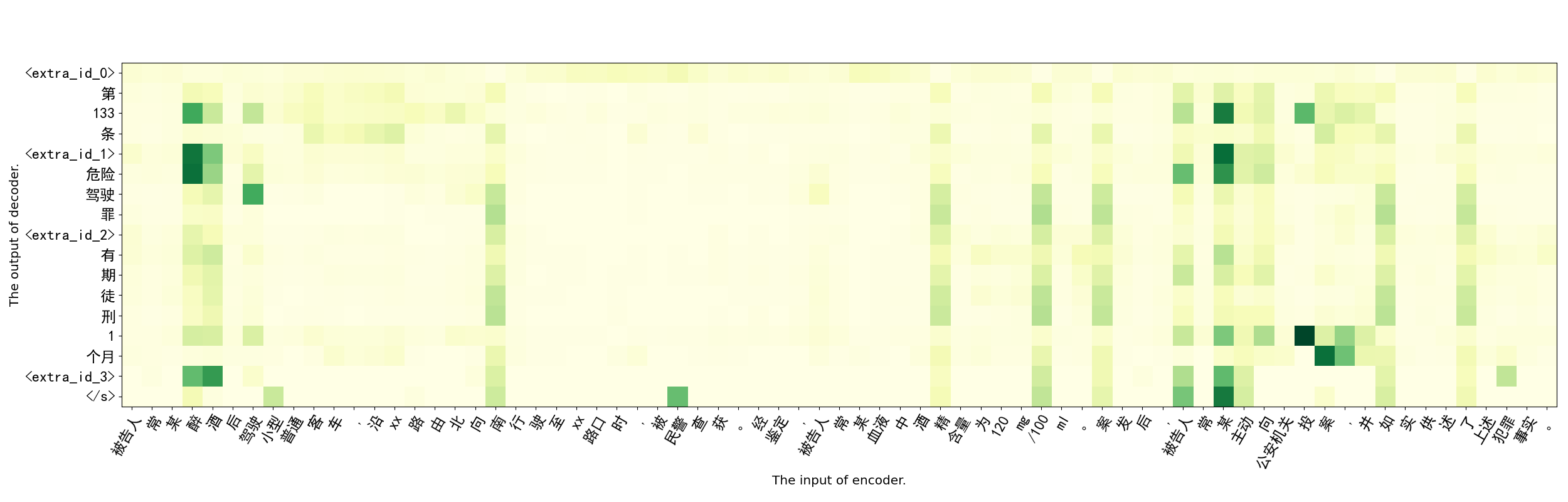}
    \includegraphics[width=.95\textwidth]{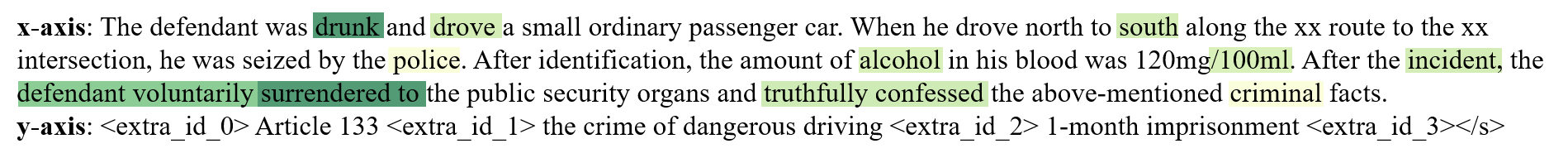}
    \centering
    \caption{\small Visualization of cross attention between the encoder and decoder. The decoder can learn to attend to proper key words in the fact to predict corresponding judgement results.}
    \label{fig_attention}
\end{figure*}

\begin{figure}[t]
\centering
\includegraphics[width=.9\linewidth]{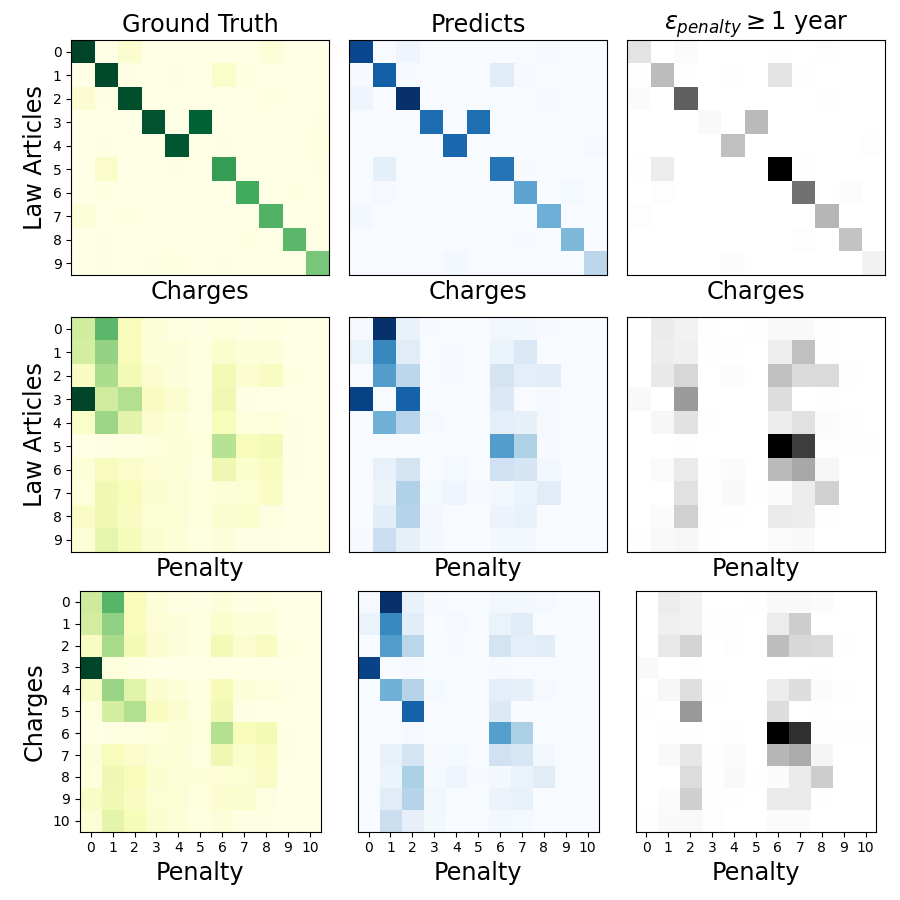}
\caption{\small Point-wise mutual information between different classes of LJP tasks. From left to right are counted on (1) ground-truth dataset, (2) results predicted by Dependent-T5 and (3) mispredicted results by Dependent-T5 with error of penalty $\geq$ 1 year.}
\label{fig4}
\end{figure}

\subsubsection{Data Size}
To observe the effect of data size on the dependency learning advantage, we trained TopJudge, STL-T5, and dependent-T5 (with the identified best decoding order $\mathcal{A} , \mathcal{C} , \mathcal{P} , \mathcal{V}$) on training sets of varying sizes. The results on test set are shown in Figure~\ref{fig:data_size}. With the increase of training data, performances of three models continue to improve. When training data grows, the gap between STL-T5 and Dependent-T5 becomes smaller. We conjecture that the effects of parameter sharing and dependency learning will saturate with sufficient data. Nonetheless, both outperform TopJudge by a large margin. Dependent-T5 is especially data efficient and achieves decent accuracy with only hundreds of training samples.

\subsubsection{Interpretability} 
Figure~\ref{fig_attention} depicts the cross attention weight distribution of each token in the input (x-axis) when generating output tokens (y-axis). Each score is obtained by summing over the attention score of each head. It shows which parts of the fact description play an important role in the generation of judgment results and also provides interpretability for the prediction.

In Figure~\ref{fig4}, we visualize the point-wise mutual information (PMI) between any two categories in LJP tasks under three scenarios: (1) ground-truth data, (2) predicted results and (3) wrongly predicted results where the predicted penalty has an error of more than 1 year. Due to the large number of categories involved in our dataset, we selected top-10 law articles and top-11 (two have equal probabilities) charges with the highest co-occurrence probabilities visualization. We also divided term of penalty into 11 categories as mentioned earlier. We can see that \emph{the PMI heatmap shares the similar shape under all three scenarios}. The model prediction follows the same correlation as the ground-truth data. Even when the model prediction makes significant errors, the correlation is still maintained. The strong correlation suggests the model has indeed learnt the proper dependency relationships, and \emph{we can use the dependent tasks to interpret the model predictions}, e.g., the wrong prediction of the following task is because of the wrong prediction of the predecessor tasks. 

\subsubsection{Case Study}
\begin{figure*}[htb]
\centering
\includegraphics[width=.95\linewidth]{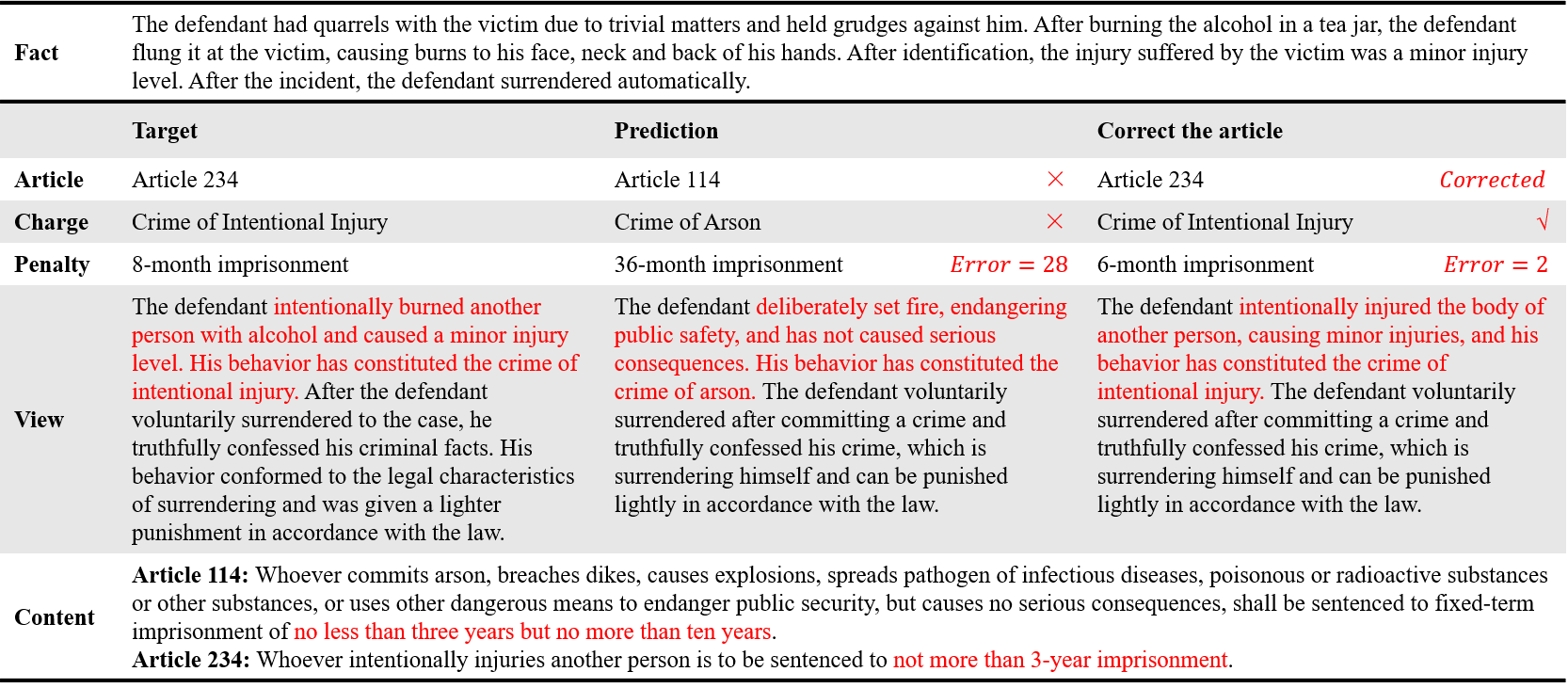}
\caption{\small A wrongly predicted example. When we replace the predicted law article with ground-truth, subsequent judgment results are corrected. The model-generated court view and law article content clearly explain why the model makes such a prediction.}
\label{fig5}
\end{figure*}

We provide an example predicted with our method in Figure~\ref{fig5}. Since the defendant causes the burn of the face, neck and hands of the victim, the model makes the wrong judgment prediction which relates the case to the crime of arson. The generated court view and relevant law article content clearly explain why the model makes such a prediction. When we replace the first decoded law article with ground-truth, results of following tasks are also corrected accordingly, and the error of predicted penalty is greatly reduced from 28 months to only 2 months. The generated court view is also changed which can provide supports for the new judgment results. Similarly, the law article content also explains why the model prediction is changed from 36 months to 6 months because the penalty of arson is stronger than the penalty of intentional injury by law. The example illustrates how we can let the model ``explain" itself by decoding the whole line of tasks, which is crucial for a transparent and trustworthy LJP system.

\section{Related Work}
Research on LJP has been studied for decades. Early researches relied primarily on hand-craft features and applied statistical~\cite{ulmer_quantitative_1963,keown_mathematical_1980} or machine learning methods~\cite{liu_exploring_2006,lin_exploiting_2012,sulea_predicting_2017,li_markov_2018} for it. Due to limitations of data, they focused on a tiny subset of case categories.

Recently, with the accessibility of large-scale judicial datasets~\cite{xiao_cail2018_2018,chalkidis_large-scale_2019,ge2021learning} and the development of deep learning~\cite{shen2017estimation,devlin2019bert,qiu2020easyaug,chang2020dart,chang2021neural,chang2021training}, rapid progress has been made by treating each sub-task in LJP as a text classification problem~\cite{jiang_interpretable_2018,he_secaps_2019}.
Some works experimented conditioning a certain sub-task on other sub-tasks and showed improvement. For example, \cite{long_automatic_2019} applied reading comprehension framework to incorporate law articles for the task of predicting judgments of civil cases. \cite{wei_external_2019} utilized given law articles as supplementary information to predict related charges. \cite{chen_charge-based_2019} proposed a deep gating network for charge-based prison term prediction. All these works rely on the ground-truth dependent sub-tasks, which are usually not available in real scenarios. When replacing the ground-truth labels with predicted ones, there is a significant drop of performance~\cite{zhong_legal_2018}.

There have been works applying multi-task learning to share model parameters among sub-tasks~\cite{zhong_iteratively_2020,xu_distinguish_2020}. As all sub-tasks are classification problems, all parameters can be shared except for the task-specific prediction heads. Nevertheless, there exists a strict logical order when humans make LJP decisions, and the multi-task framework lacks the mechanism to reflect this ordered dependency. Therefore, some researchers try to establish explicit ordered dependencies among sub-tasks. For example, after encoding facts with neural network, \cite{luo_learning_2017} extracted top k law articles with SVM and utilized them to assist the prediction of charges. \cite{zhong_legal_2018, yang_legal_2019} built dependencies among all three sub-tasks. They apply CNN for fact encoding and use LSTM to model the ordered dependencies. Each sub-task gets a task-specific representation to predict the result. In comparison, our model leverages a single unified architecture that can naturally build dependencies for arbitrary types of sub-tasks.

There have also been works focusing on court view generation~\cite{ye_interpretable_2018,wu2020biased,li2021court}, but did not leverage it to improve the LJP results. Some works utilized law article contents to assist charge prediction~\cite{wang2018modeling, yang2019recurrent}. They design complex architectures to incorporate the contents. In contrast, our framework is simple and \emph{does not affect the inference speed at all} since the content is only used for training.

Leveraging a single unified text-to-text Transformer has also been applied in other NLP tasks like dialogue generation~\cite{hosseini2020simple,su2020moviechats} and question answering~\cite{oguz2020unified}. We adopt a similar approach in our work and further show its flexibility of enabling effective dependency learning.

\section{Ethical Statement}
Since LJP is an emerging but sensitive technology, we would like to discuss ethical concerns of our work. Firstly, the corpus is created from publicly available data and the personal private information (e.g., name, plate number, etc.) has been anonymized. In addition, the proposed method aims to assist legal professionals in their research and decision-making instead of replacing them. Therefore, ethical considerations such as allowing legal rights and obligations of human beings to be decided by non-human intelligence are not breached by the system.

\section{Conclusion}
In this paper, we propose learning dependencies among sub-tasks of LJP with a unified text-to-text Transformer. It directly leverages the original legal judgment document as the prompt without handcrafting, enables full parameter sharing and supports arbitrary types and amounts of sub-tasks. We show it significantly outperforms SOTA, and the model can learn the task dependencies more effectively than previous methods. We analyze the effects of decoding order, data size and illustrate it can provide explanations to its own prediction results, even when an error is made. The proposed framework is simple and flexible. It can be easily extended to cover more legal tasks, which we leave for future work.


%



\ifCLASSOPTIONcompsoc
  \section*{Acknowledgments}
\else
  \section*{Acknowledgment}
\fi

This work was supported by the National Key R\&D Program of 
China (2016YFC0800803 ). 
Chuanyi Li is the corresponding author.

\ifCLASSOPTIONcaptionsoff
  \newpage
\fi



\bibliographystyle{IEEEtran}
\bibliography{ref.bib}
\end{document}